# MAP Estimation, Message Passing, and Perfect Graphs

**Tony Jebara**
Columbia University
New York, NY 10027
jebara@cs.columbia.edu

## Abstract

Efficiently finding the maximum a posteriori (MAP) configuration of a graphical model is an important problem which is often implemented using message passing algorithms. The optimality of such algorithms is only well established for singly-connected graphs and other limited settings. This article extends the set of graphs where MAP estimation is in P and where message passing recovers the exact solution to so-called perfect graphs. This result leverages recent progress in defining perfect graphs (the strong perfect graph theorem), linear programming relaxations of MAP estimation and recent convergent message passing schemes. The article converts graphical models into nand Markov random fields which are straightforward to relax into linear programs. Therein, integrality can be established in general by testing for graph perfection. This perfection test is performed efficiently using a polynomial time algorithm. Alternatively, known decomposition tools from perfect graph theory may be used to prove perfection for certain families of graphs. Thus, a general graph framework is provided for determining when MAP estimation in any graphical model is in P, has an integral linear programming relaxation and is exactly recoverable by message passing.

## 1 INTRODUCTION

Recovering the maximum a posteriori (MAP) configuration of random variables in a graphical model is an important problem with applications ranging from protein folding to image processing. Graphical models (which include Bayesian networks and Markov random fields) use a graph over dependent random variables to compactly express a probability density function as a product of functions over maximal cliques in the graph. For a general graphical model, the MAP problem is NP-hard (Shimony, 1994). A popular algorithm for approximating the MAP solution is max-product belief propagation and its variants (Weiss, 2000; Globerson & Jaakkola, 2007) which operate by sending messages between neighboring cliques until convergence. It is known that max-product belief propagation converges to the optimum on singly-linked graphs and junction-trees (Pearl, 1988; Wainwright & Jordan, 2008). Less is known about its formal properties when graphs contain loops. In practice, however, there are multiple applied cases in the literature where the the max-product algorithm performs extremely well on graphs with loops. For example, turbo codes, one of the top performing error correcting coding schemes to date, can be successfully implemented via max-product on a loopy graph (Weiss & Freeman, 2001). Recently, formal guarantees for such algorithms have been found for graphs with a single loop (Weiss, 2000), maximum weight bipartite matching graphs (Bayati et al., 2005), and maximum weight bipartite $b$-matching graphs (Huang & Jebara, 2007). In these settings, since the graphs contain loops, message passing algorithms are often referred to as loopy belief propagation or loopy message passing. While the single loop case is of limited practical use, the matching and $b$-matching message passing algorithms have many applications and lead to competitive methods for solving large-scale matching problems. Subsequently, additional results for matching and $b$-matching problems (Sanghavi et al., 2008; Bayati et al., 2008) were produced by examining the linear program (LP) relaxation (Santos, 1991; Wainwright et al., 2005; Weiss et al., 2007) of the integer problem being solved during MAP estimation. Loosely speaking, if the LP relaxation of the matching problem has an integral solution, message passing converges to the MAP solution. In principle, this extends convergence arguments for matching from bipartite settings to *some* unipartite settings if the LP relaxation has integral solution. Of course, matchings and *b*-matchings are exactly solvable for both the bipartite and the more general unipartite case in polynomial time using Edmonds' Blossom algorithm (Edmonds, 1965). However, belief propagation methods may be faster and, under mild assumptions, find maximum weight matchings in no more than $O(n^2)$ time (Salez & Shah, 2009).

This article will identify general conditions on loopy graphical models such that a) the LP relaxation is integral and b) message passing will always converge to the MAP solution[1]. This extends the current list of graphical models where MAP is known to be efficient (and message passing is known to be exact) to the broader family of *perfect graphs*. Perfect graphs subsume trees, bipartite matchings and *b*-matchings and lead to a generalization of König's theorem: the so-called *weak perfect graph theorem* which states that a graph is perfect if and only if its complement is perfect (Lovász, 1972). Recently, a further generalization was proved: the *strong perfect graph theorem* which states that all perfect graphs are Berge graphs (Chudnovsky et al., 2006). Furthermore, a polynomial time algorithm was identified that verifies if a graph is perfect or not (Chudnovsky et al., 2005). To exploit these results from the combinatorics community, this article converts any graphical model into an alternative form referred to as a nand Markov random field. Therein, the integrality of the LP relaxation can be easily verified by recognizing perfect graphs. This makes it possible to precisely characterize which loopy graphs have the appropriate topology for exact MAP estimation via linear programming or message passing.

This article is organized as follows. Section 2 describes the factorization properties of graphical models and Section 3 shows a conversion into an equivalent form called a nand Markov random field (NMRF). Section 4 shows how the LP relaxation of the NMRF produces a so-called set packing linear program whose integrality properties are well characterized by the perfection of the associated graph. Section 5 defines perfect graphs as well as discusses tools and polynomial time algorithms for recognizing perfect graphs. Perfection ensures that the LP is integral and achieves the MAP estimate. Section 6 provides some pruning procedures for NMRF graphs that can make perfection tests and LP solutions more widely applicable. Section 7 uses message passing as a faster alternative to linear programming for obtaining the MAP estimate. We conclude with experiments and a brief discussion.

## 2 GRAPHICAL MODELS

A graphical model is an undirected graph used to represent the factorization of a probability distribution (Wainwright & Jordan, 2008). Consider an undirected graph $G = (V, E)$ with vertices $V = \{v_1, \ldots, v_n\}$ and edges $E : V \times V \to \mathbb{B}$. Denote the set of vertices by $\mathsf{V}(G)$ and the neighbors of a node $v_i$ by $\mathsf{Ne}(v_i)$ or $\mathsf{Ne}(i)$. The graph $G$ describes the dependencies between a set of random variables $X = \{x_1, \ldots, x_n\}$ where each variable $x_i$ is associated with a vertex $v_i$ in the graph[2]. We will assume that each $x_i \in \mathbb{Z}$ is a discrete variable[3] with $|x_i|$ settings[4]. For example, if $x_i$ is a binary variable, $0 \leq x_i < 2$ and $|x_i| = 2$. A graphical model describes a probability density over all random variables $p(X)$ which obeys the following factorization:

$$p(X) = \frac{1}{Z} \prod_{c \in C} \psi_c(X_c) \quad (1)$$

where $Z$ is a partition function (a scalar that normalizes the density), $C$ is the set of maximal cliques in the graph $C \subseteq G$ and $\psi_c(X_c)$ are positive compatibility functions over variables in each clique $c$. In other words, $X_c = \{x_i | i \in c\}$. Clearly, this representation of $p(X)$ is exponential in the cardinality of the cliques[5]. Without loss of generality, this article assumes all $\psi_c(X_c)$ are uniformly scaled such that $\psi_c(X_c) > 1$ (and $Z$ is scaled appropriately for normalization) as follows:

$$\psi_c(X_c) \leftarrow \frac{\psi_c(X_c)}{\min_{X_c} \psi_c(X_c)} + \epsilon$$

where $\epsilon$ is an infinitesimal quantity.

It is possible to convert Equation 1 into an equivalent pairwise Markov random field (MRF) over binary variables (Ravikumar & Lafferty, 2006; Yedidia et al., 2001) at the expense of increasing the state space. Section 3 follows such an approach but restricts the conversion further by requiring that all potential functions enforce nand relationships among binary variables.

## 3 NAND MARKOV RANDOM FIELDS

Any generic graphical model with graph $G$ in Equation 1 can be converted into an equivalent graphical model with graph $\mathcal{G}$ which will be referred to as a nand Markov random field (NMRF). In this form, all clique functions involve a nand operation over binary variables as $\psi_c(X_c) = \delta(\sum_{x \in X_c} x \leq 1)$ where we take the function $\delta \in \mathbb{B}$ to equal 1 if the statement inside is true and 0 otherwise. Clearly, these clique functions factorize into a product over pairwise edges since $\psi_c(X_c) = \prod_{x_i \neq x_j \in X_c} \delta(x_i + x_j \leq 1)$. Indeed, graphical models for solving maximum weight matchings are usually in this form (Sanghavi et al., 2008; Bayati et al., 2008). The NMRF form helps produce linear programming relaxations of the MAP problem which have desirable properties as detailed in Section 4.

---

[1]For brevity, this article does not discuss cases where multiple MAP solutions (i.e. ties) are possible. Any such solution is assumed acceptable as a MAP estimate.

[2]In this article, the variable $x_i$, the node $v_i$ and the index $i$ will be used interchangeably when the meaning is evident from the context.

[3]While graphical models can handle cases where $x_i$ are scalars, this article only deals with discrete $x_i$.

[4]Here, $|\cdot|$ is the cardinality of a variable or a set.

[5]This article focuses on polynomial efficiency in the number of cliques and will not address problems related to exponential dependence on maximum clique size.

Consider forming an NMRF from $G$ which represents a distribution over a set $\mathbf{X}$ of $N$ binary variables $\mathbf{x} \in \mathbb{B}$. For each clique $c \in C$ in the original graph $G$, introduce binary variables $\mathbf{x}_{c,k}$ for each *configuration* of the arguments of the clique function $\psi_c(X_c)$. In other words, for clique $X_c$, define a set of binary variables $\mathbf{X}_c = \{\mathbf{x}_{c,1}, \ldots, \mathbf{x}_{c,|\mathbf{X}_c|}\}$ with $|\mathbf{X}_c| = \prod_{i \in c} |x_i|$. The NMRF represents a distribution over all such variables $\mathbf{X} = \cup_{c \in C} \mathbf{X}_c$ and, since all $\mathbf{X}_c$ are disjoint (with redundant instantiations of the variables in each clique $X_c$), the state space of the NMRF has cardinality

$$|\mathbf{X}| = \sum_{c \in C} \left( \prod_{i \in c} |x_i| \right) = N. \tag{2}$$

Given a setting of $X = \{x_1, \ldots, x_n\}$, the corresponding setting of $\mathbf{X} = \{\mathbf{x}_1, \ldots, \mathbf{x}_N\}$ is given by:

$$\mathbf{x}_{c,k} = \begin{cases} 1 & \text{if } k = 1 + \sum_{i \in c} x_i \left( \prod_{j=1}^{i-1} |x_i|^{\delta(j \in c)} \right) \\ 0 & \text{otherwise.} \end{cases} \tag{3}$$

This is a mapping from $X$ to a setting of $\mathbf{X}$ as an injection since some settings of $\mathbf{X}$ yield invalid settings of $X$ if they involve disagreement in the clique configurations. The expression (which is admittedly inelegant) merely states that when $X_c$ is in its $k^{\text{th}}$ configuration from among its total of $\prod_{i \in c} |x_i|$ possible configurations, we must have $\mathbf{x}_{c,k} = 1$ in the NMRF.

It is now possible to write an equivalent function $\rho(\mathbf{X})$ which mimics Equation 1. This need not be a normalized probability density function over the space $\mathbf{X}$ since we are only interested in its maximization for the MAP estimate. The function $\rho(\mathbf{X})$ is as follows

$$\rho(\mathbf{X}) = \prod_{c \in C} \Psi_c(\mathbf{X}_c) \prod_{k=1}^{|\mathbf{X}_c|} e^{f_{c,k} \mathbf{x}_{c,k}} \prod_{\substack{d \in C \\ d \neq c}} \prod_{l=1}^{|\mathbf{X}_d|} \Phi(\mathbf{x}_{c,k}, \mathbf{x}_{d,l})^{\mathbf{z}_{c,k,d,l}} \tag{4}$$

where, once again, $C$ is the set of maximal cliques in the graph $C \subseteq G$ and $\Psi_c(\mathbf{X}_c)$ are compatibility functions over sets of binary variables. Furthermore, $\mathbf{z}_{c,k,d,l}$ variables are binary switches to be defined subsequently. To mimic the original $p(X)$, the factorization contains a product over $\exp(f_{c,k} \mathbf{x}_{c,k})$ involving non-negative scalars

$$f_{c,k} = \log \psi_c(X_c)$$

where the appropriate configuration for $X_c$ is recovered from $(c, k)$ as determined by the relationship in Equation 3. Note that all $f_{c,k} > 0$ since $\psi_c(X_c) > 1$. Finally, the factorization contains additional potential functions $\Phi(\mathbf{x}_{c,k}, \mathbf{x}_{d,l})$ for each pair of variables $\mathbf{x}_{c,k}$ and $\mathbf{x}_{d,l}$ if the binary variable $\mathbf{z}_{c,k,d,l}$ equals unity (otherwise, the functions are taken to the power of 0 and disappear from the product). The important difference with this model and the one in Equation 1 is that all its non-singleton clique potential functions $\Psi_c(\mathbf{X}_c)$ and pairwise functions $\Phi(\mathbf{x}_{c,k}, \mathbf{x}_{d,l})$ accept binary values and produce binary outputs as nand operations:

$$\Psi_c(\mathbf{X}_c) = \begin{cases} 1 & \text{if } \sum_k \mathbf{x}_{c,k} \leq 1 \\ 0 & \text{otherwise} \end{cases}$$

$$\Phi(\mathbf{x}_{c,k}, \mathbf{x}_{d,l}) = \begin{cases} 1 & \text{if } \mathbf{x}_{c,k} + \mathbf{x}_{d,l} \leq 1 \\ 0 & \text{otherwise.} \end{cases}$$

The binary variable $\mathbf{z}_{c,k,d,l}$ indicates a potential disagreement between $\mathbf{x}_{c,k}$ and $\mathbf{x}_{d,l}$ over settings of the variables in $X_c \cap X_d$ that they are both jointly implicated in. This is defined more formally as follows:

$$\mathbf{z}_{c,k,d,l} = 1 - \prod_{i=1}^n \delta \left( \mathrm{mod}\left( \left\lfloor \frac{k-1}{\prod_{j=1}^{i-1} |x_j|^{\delta(j \in c)}} \right\rfloor, |x_i| \right) = \right.$$
$$\left. \mathrm{mod}\left( \left\lfloor \frac{l-1}{\prod_{j=1}^{i-1} |x_j|^{\delta(j \in d)}} \right\rfloor, |x_i| \right) \right)^{\delta(i \in c) \delta(i \in d)}$$

where it is understood that $0^0 = 1$.

It is now straightforward to consider the undirected graph $\mathcal{G} = (\mathcal{V}, \mathcal{E})$ implied by Equation 4 which is recovered from the original graph $G = (V, E)$. This graph contains nodes $\mathcal{V} = \{\mathbf{v}_{c,k} : \forall c \in C, k = 1, \ldots, |\mathbf{X}_c|\}$ where each node $\mathbf{v}_{c,k}$ is associated with a corresponding variable $\mathbf{x}_{c,k}$. The graph $\mathcal{G}$ then has edges between all pairs of nodes $\mathbf{v}_{c,k}$ corresponding to variables in the clique $\mathbf{X}_c$ for $c \in C$. Furthermore, for all pairs of nodes $\mathbf{v}_{c,k}$ and $\mathbf{v}_{d,l}$ are connected if $\mathbf{z}_{c,k,d,l} = 1$. The formula for the set of edges in $\mathcal{G}$ simplifies as:

$$\mathcal{E}(\mathbf{v}_{c,k}, \mathbf{v}_{d,l}) = \max\left(\delta(c = d)\delta(k \neq l), \mathbf{z}_{c,k,d,l}\right)$$
$$= \mathbf{z}_{c,k,d,l}.$$

This results in an undirected graph $\mathcal{G}$ of pairwise nand functions and Equation 4 can be written as:

$$\rho(\mathbf{X}) = \prod_{c \in C} \prod_{k=1}^{|\mathbf{X}_c|} e^{f_{c,k} \mathbf{x}_{c,k}} \prod_{d \in C} \prod_{l=1}^{|\mathbf{X}_d|} \Phi(\mathbf{x}_{c,k}, \mathbf{x}_{d,l})^{\mathbf{z}_{c,k,d,l}}$$

although Equation 4 more clearly distinguishes between intra-clique edges arising from $X_c$ and inter-clique edges arising from $X_c \cap X_d$. Thus, the NMRF contains nand edges between all pairs of binary variables that cannot be jointly instantiated without causing a disagreement. For each edge, only one or fewer of the vertices adjacent[6] to it may be instantiated (equal to unity); hence the term nand Markov random field. For instance, the functions $\Psi_c(\mathbf{X}_c)$ place edges between all variables corresponding to differing configurations of $X_c$, at most one of which may be active (i.e. equal to one) at any time. Thus, all the potential functions in this graphical model are acting as nand gates and all edges in the graph enforce a nand relationship between the nodes they are adjacent to. This graphical model is reminiscent of the MRF used in (Ravikumar & Lafferty, 2006) which had xor potential functions requiring that the variables inside cliques sum strictly to 1. The NMRF, on the other hand, requires a nand relationship: pairs of variables sum to $\leq 1$. Figure 1 displays a graphical model and its corresponding NMRF.

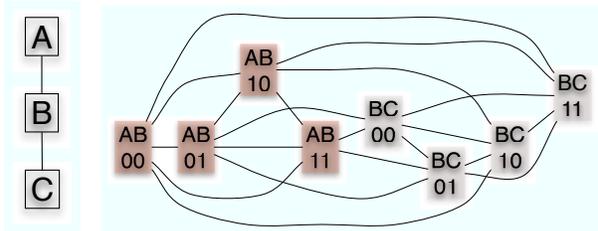

Figure 1: A graphical model (left) over binary variables with cliques $\{A, B\}$ and $\{B, C\}$ and its corresponding binary nand Markov random field (right).

It remains to show that the MAP estimate $\mathbf{X}^*$ of $\rho(\mathbf{X})$ corresponds to a valid MAP estimate $X^*$ of $p(X)$ despite the surjective relationship between $X$ and $\mathbf{X}$. Since the variables in $\mathbf{X}$ correspond to possibly disagreeing settings of $X$, only some binary configurations are admissible in $\mathbf{X}$. This is because every clique $X_c$ must be in one configuration and overlapping cliques may not disagree in their configurations. However, the constraints in Equation 4 only require $\sum_k \mathbf{x}_{c,k} \leq 1$. This permits the possibility that some cliques will simply not be assigned a configuration when the MAP estimate is recovered from Equation 4. In other words, it may be the case that $\sum_k \mathbf{x}_{c,k} = 0$. The next theorem shows that the MAP estimate $\mathbf{X}^*$ will always produce $\sum_k \mathbf{x}_{c,k} = 1$ for all $c \in C$.

**Theorem 1** *Given the maximum a posteriori estimate* $\mathbf{X}^* = \{\mathbf{x}_1^*, \dots, \mathbf{x}_{|\mathbf{X}|}^*\}$ *of Equation 4, all variables in cliques* $c \in C$ *satisfy* $\sum_k \mathbf{x}_{c,k}^* = 1$.

---

[6]Adjacent vertices are vertices connected by an edge.

**Proof 1** *The MAP solution involves binary settings* $\mathbf{x}_{c,k}^* \in \{0, 1\}$ *for all variables in* $\mathbf{X}^*$. *Setting* $\mathbf{X}$ *to all zeros produces a value* $\rho(\mathbf{X}) = 1$ *since all functions* $\Psi$ *and* $\Phi$ *are satisfied and all the values of* $f_{c,k}$ *are multiplied by zero prior to exponentiation. Therefore, assume that the maximizer is not the all-zeros configuration and that* $\rho(\mathbf{X}^*) > 1$, *since, otherwise, all settings of $X$ trivially produce a MAP estimate. Requiring* $\rho(\mathbf{X}^*) > 1$ *corresponds to having at least one nonzero setting in* $\mathbf{X}^*$. *Choose this binary variable as* $\mathbf{x}_{\hat{c},\hat{k}}^* = 1$ *which now satisfies* $\sum_k \mathbf{x}_{\hat{c},k}^* = 1$ *to produce* $\rho(\mathbf{X}^*) > 1$ *since* $f_{c,k} > 0$ *and* $\exp(f_{c,k} \mathbf{x}_{c,k}) > 1$. *Subsequently, there can be no disagreement between the configurations of overlapping cliques since pairwise potential functions* $\Phi(\mathbf{x}_{c,k}, \mathbf{x}_{d,l})$ *exist between all pairs of binary variables when* $\mathbf{z}_{c,k,d,l} = 1$ *and setting binary variables corresponding to conflicting assignments for $X_c$ and $X_d$ would force* $\rho(\mathbf{X}) = 0$. *Thus, there can be no disagreement in the configurations of the cliques. If* $\rho(\mathbf{X}^*) > 1$, *it must be the case that either of the following holds:* $\sum_k \mathbf{x}_{c,k} = 1$ *or* $\sum_k \mathbf{x}_{c,k} = 0$ *for each* $c \neq \hat{c}$. *Consider finding a clique* $\tilde{c} \in C \setminus \hat{c}$ *where the latter case is true. There,* $\tilde{c}$ *has no assigned configuration for its variables* $\mathbf{X}_{\tilde{c}}$ *and* $\sum_k \mathbf{x}_{\tilde{c},k} = 0$. *For any such clique* $\tilde{c}$ *there is always a configuration that may be selected which agrees with neighboring cliques. Since every value of* $f_{\tilde{c},k} > 0$, *it is always possible to preserve agreement and set one of the* $\mathbf{x}_{\tilde{c},k}$ *to unity to strictly increase* $\rho(\mathbf{X})$ *while preserving agreement. Repeating this line of reasoning on all remaining cliques only further increases* $\rho(\mathbf{X})$ *until all cliques satisfy* $\sum_k \mathbf{x}_{c,k} = 1$. *Thus, the NMRF produces a MAP estimate satisfying* $\sum_k \mathbf{x}_{c,k}^* = 1$ *for all cliques* $c \in C$.

**Lemma 1** *The MAP estimate of Equation 4 corresponds to the MAP estimate of Equation 1.*

**Proof 2** *Since all configurations are in agreement and* $\sum_k \mathbf{x}_{c,k} = 1$, *the maximizer* $\mathbf{X}^*$ *of Equation 4 corresponds to a valid setting of $X^*$ and we can associate $X^*$ with* $\mathbf{X}^*$. *It is straightforward to see that* $\rho(\mathbf{X}^*)/Z = p(X^*)$. *Since* $\rho(\mathbf{X}^*) \geq \rho(\mathbf{X})$ *for all $\mathbf{X}$ and $\mathbf{X}$ spans a strict superset of the configurations of $X$, it must be the case that* $p(X^*) \geq p(X)$ *for all $X$.*

The next section will show that, when $\mathcal{G}$ corresponds to a perfect graph, the LP relaxation of Equation 4 is integral. In those settings the MAP estimate can be recovered by linear programming.

## 4 PACKING LINEAR PROGRAMS

Consider the LP relaxation of the MAP estimation problem on the NMRF in Equation 4 (which was shown to be equivalent to MAP estimation with the graphical model in Equation 1). A linear program is an optimization over a vector of variables $\vec{x} \in \mathbb{R}^N$ which are used as surrogates for the binary variables $\mathbf{X}$ in the MAP problem on the NMRF. If the LP is tight and gives back an integral solution, then $\vec{x}$ recovers the exact MAP estimate. Denote the all-ones vector $\vec{1} \in \mathbb{R}^N$. In general, linear programming (or any convex optimization problem) can be solved in time cubic in the number of variables. The following theorem strictly characterizes when an LP in known specifically as a *set packing linear program* (which explores the set packing polytope) yields integral solutions $\vec{x}^* \in \{0,1\}^N$.

**Theorem 2** *(Lovász, 1972; Chvátal, 1975) For every non-negative vector $\vec{f} \in \mathbb{R}^N$, the linear program*

$$\beta = \max_{\vec{x} \in \mathbb{R}^N} \vec{f}^\top \vec{x} \text{ subject to } \vec{x} \geq 0 \text{ and } A\vec{x} \leq \vec{1}$$

*recovers a vector $\vec{x}$ which is integral if and only if the (undominated) rows of $A$ form the vertex versus maximal cliques incidence matrix of some perfect graph.*

We say the $d^{\text{th}}$ row of a matrix $A$ is undominated if there is no row index $c \neq d$ such that $A_{cj} \leq A_{dj}$ for all $j = 1, \ldots, N$. Let $\mathcal{G}$ be a graph with vertices $\mathcal{V} = \{\mathbf{v}_1, \mathbf{v}_2, \ldots, \mathbf{v}_N\}$ and $\{\mathbf{V}_1, \ldots, \mathbf{V}_{|\mathcal{C}|}\}$ its (inclusion-wise) maximal cliques. We define the incidence matrix of $\mathcal{G}$ as $A \in \mathbb{B}^{|\mathcal{C}| \times N}$ where $A_{cj} = 1$ if $\mathbf{v}_j \in \mathbf{V}_c$ and $A_{cj} = 0$ otherwise.

Theorem 2 describes when the above LP will yield an integer solution. For general graphs $G$ and general Markov random fields $\mathcal{G}$, the MAP estimate is NP (Shimony, 1994). Remarkably, by examining the topology of the graph $\mathcal{G}$, it is possible to characterize exactly when the linear programming relaxation will be integral (or otherwise) for any NMRF $\mathcal{G}$. If the graph $\mathcal{G}$ is a *perfect* graph, then its LP relaxation is integral and the MAP estimate can be recovered in polynomial (at most cubic) time. This is summarized in the following theorem.

**Theorem 3** *The MAP estimate of the nand Markov random field in Equation 4 is in P if the graph $\mathcal{G}$ is perfect and MAP estimation takes at most $O(|V(\mathcal{G})|^3)$ by linear programming if $\mathcal{G}$ is perfect.*

**Proof 3** *The LP relaxation of the MAP estimate of the nand Markov random field produces a packing linear program. Given the graph $\mathcal{G}$, it is straightforward to recover its corresponding vertex versus maximal cliques incidence matrix $A$. Taking the logarithm of Equation 4 shows that the MAP optimization is exactly equivalent to the LP in Theorem 2. The LP is a direct relaxation of the binary variables in Equation 4 and the matrix $A$ corresponds to the graph $\mathcal{G}$, the vector $\vec{x} = \text{vec}(\mathbf{X})$ is the concatenation of all the binary random variables and the vector $\vec{f}$ is defined element-wise as the logarithm of the clique functions for every clique and every configuration:*

$$\vec{f} = [\log(\psi_c(X_c)) : \forall c \in C, \forall X_c]^\top.$$

*Recall that $\log(\psi(X_c))$ is always positive since all clique potential functions satisfy $\psi(X_c) > 1$ in the original graph $G$. Therefore, Equation 4 corresponds to the LP in Theorem 2. If $\mathcal{G}$ is a perfect graph, the integrality of the LP is established via Theorem 2 and linear programming achieves the MAP estimate.*

The essential test is to show that $\mathcal{G}$ is (or is not) a perfect graph, which, in turn, determines conclusively if the LP is (or is not) integral. It is then possible to relate the result on the NMRF above to general graphical models via the following corollary.

**Corollary 1** *The MAP estimate of any graphical model with cliques $c \in C$ over variables $\{x_1, \ldots, x_n\}$ that produces a nand Markov random field as in Equation 4 with a perfect graph $\mathcal{G}$ is in P and can be computed in at most $\mathcal{O}\left(\left(\sum_{c \in C} \left(\prod_{i \in c} |x_i|\right)\right)^3\right)$.*

**Proof 4** *Theorem 1 ensures that the MAP estimate of the nand Markov random field produces the MAP estimate of the graphical model. Theorem 3 shows that recovering the MAP estimate of the NMRF is in P and is cubic in the number of vertices. The number of vertices of the NMRF is given by Equation 2.*

In summary, if graph $\mathcal{G}$ is a perfect graph, the LP relaxation is integral and recovers the MAP estimate of the NMRF in Equation 4 as well as the MAP estimate of the graphical model in Equation 1. Linear programming is cubic in the number of variables. However, Section 7 discusses message passing algorithms which often yield better efficiency in practice. First, however, we discuss perfect graphs and their construction and, in particular, a polynomial time algorithm that answers if a graph is perfect or is not.

## 5 PERFECT GRAPHS

A perfect graph (Berge, 1963; Lovász, 1983) is a graph where every induced subgraph has chromatic number equal to its clique number. The clique number of a graph $\mathcal{G}$ is denoted $\omega(\mathcal{G})$ and is the size of the maximum clique (fully connected subgraph) of $\mathcal{G}$. The chromatic number of $\mathcal{G}$, $\chi(\mathcal{G})$, is the minimum number of colors needed to label vertices such that no two adjacent vertices have the same color. Perfect graphs have the remarkable property, $\omega(\mathcal{H}) = \chi(\mathcal{H})$ for every induced subgraph $\mathcal{H} \subseteq \mathcal{G}$. Perfect graphs also have computational properties (Grötschel et al., 1988). For instance, in all perfect graphs, potentially intractable problems such as graph coloring, maximum clique and maximum independent set are in P.

In recent work (Chudnovsky et al., 2006), the strong perfect graph conjecture as described in (Berge, 1963; Berge & Ramírez-Alfonsín, 2001) was proved. Namely, a graph is perfect if an only if it is Berge. A Berge graph is a graph that contains no odd hole and whose complement also contains no odd hole; both terms are defined below.

**Definition 1 (Graph Complement)** *The complement $\bar{\mathcal{G}}$ of a graph $\mathcal{G}$ is a graph with the same vertex set $V(\mathcal{G})$ as $\mathcal{G}$, where distinct vertices $\mathbf{u}, \mathbf{v} \in V(\mathcal{G})$ are adjacent in $\bar{\mathcal{G}}$ just when they are not adjacent in $\mathcal{G}$. The complement of the complement of a graph gives back the original graph.*

**Definition 2 (Hole)** *A hole of a graph $\mathcal{G}$ is an induced subgraph of $\mathcal{G}$ which is a chordless cycle of length at least $5$. An odd (even) hole is a chordless cycle with odd (even) length.*

The proof of the strong perfect graph conjecture (Chudnovsky et al., 2006) conclusively showed that a graph is perfect if and only if it is a Berge graph. The proof also specifies that any Berge graph must belong to one of the following basic classes of Berge graph:

- bipartite graphs
- complements of bipartite graphs
- line graphs of bipartite graphs
- complements of line graphs of bipartite graphs
- double split graphs

or admit one of four structural decompositions:

- a 2-join
- a 2-join in the complement
- an $M$-join
- a balanced skew partition.

These decompositions are ways of breaking up the graph such that the remaining parts may eventually be recognized as basic Berge graphs. Note, a line graph $\mathsf{L}(\mathcal{G})$ of a graph $\mathcal{G}$ is a graph which contains a vertex for each edge of $\mathcal{G}$ and where two vertices of $\mathsf{L}(\mathcal{G})$ are adjacent if and only if they correspond to two edges of $\mathcal{G}$ with a common end vertex.

The family of perfect graphs makes it possible to precisely characterize if a graphical model $G$ (or more precisely, its equivalent nand Markov random field $\mathcal{G}$) admits efficient MAP estimation. Also, remarkably, automatically *verifying* if *any* graph is perfect is efficient. Recently, a polynomial time algorithm (in the number of vertices of the graph) was introduced to test if a graph is perfect.

**Theorem 4** *(Chudnovsky et al., 2005) Determining if graph $\mathcal{G}$ is perfect is P and takes at most $\mathrm{O}(|V(\mathcal{G})|^9)$.*

Given a graph $\mathcal{G}$, the algorithm decides either that $\mathcal{G}$ is not Berge or that $\mathcal{G}$ contains no odd hole. To test Bergeness, the algorithm is run on $\mathcal{G}$ and again on $\bar{\mathcal{G}}$. The key computational bottleneck is the detection of so-called pyramid structures by enumerating all nonuples (leading to a ninth order polynomial run-time) of vertices and considering various shortest paths between them. Further details of the algorithm are omitted in this article for space considerations but implementation is straightforward. This polynomial time algorithm leads to the following straightforward corollary for graphical models (via the conversion to NMRFs).

**Corollary 2** *Verifying if MAP estimation is efficient for any graphical model with cliques $c \in C$ over variables $\{x_1, \ldots, x_n\}$ is in P and takes at most $\mathrm{O}\left(\left(\sum_{c \in C} \left(\prod_{i \in c} |x_i|\right)\right)^9\right)$ time.*

Therefore, an automatic framework is possible for verifying if MAP estimation of any graphical model is in P. The model is first converted into a nand Markov random field with a graph $\mathcal{G} = (\mathcal{V}, \mathcal{E})$ and then the resulting graph is efficiently tested using the algorithm of (Chudnovsky et al., 2005). If the resulting graph is perfect, the LP relaxation efficiently recovers the MAP estimate. Unfortunately, the current running time of the perfect graph verification algorithm prohibits practical application. Only small graphical models $G$ can be efficiently tested to date: those that map to a corresponding NMRF graph $\mathcal{G}$ with less than 20 nodes. It may be helpful to consider the faster heuristic algorithm of (Nikolopoulos & Palios, 2004) which only requires $\mathrm{O}(|\mathcal{V}| + |\mathcal{E}|^2)$. This algorithm only verifies if a graph contains any hole or chordless cycle with 5 or more nodes. Thus, if the graph and its complement contain no holes (even or odd), the algorithm can quickly confirm that $\mathcal{G}$ is perfect. However, if the graph contains holes, it is still unclear whether these are exclusively even holes or if there are some odd holes in the graph. Therefore, (Chudnovsky et al., 2005) becomes necessary as the conclusive test for graph perfection.

Clearly, the above algorithms may be impractical for large scale problems. Fortunately, a variety of decomposition and construction tools are also available from perfect graph theory which may be useful to formally prove perfection without cumbersome computation. These include the replication lemma (Lovász, 1972), the 2-join decomposition theorem (Cornuéjols & Cunningham, 2001), the $M$-join decomposition and the skew-partition decomposition theorem (Chudnovsky et al., 2006). In the remainder of this section, a direct proof approach is used to investigate popular graphical models where MAP estimation is known to be easy to see if these indeed produce NMRFs with perfect graphs.

Consider the following tool from perfect graph theory known as the replication lemma.

**Lemma 2** *(Lovász, 1972) Let $\mathcal{G}$ be a perfect graph and let $v \in V(\mathcal{G})$. Define a graph $\mathcal{G}'$ by adding a new vertex $v'$ and joining it to $v$ and all the neighbors of $v$. Then $\mathcal{G}'$ is perfect.*

This tool will be useful for investigating graphical models where $G$ is a tree (Pearl, 1988).

**Lemma 3** *A graphical model with a tree graph $G$ produces an NMRF with a perfect graph $\mathcal{G}$.*

**Proof 5** *First consider the simplest case where the input tree graph is merely a star graph. A star graph $G_v$ consists of leaf nodes $\{v_1, \ldots, v_{|C|}\}$, a single internal node $v$ present in a total of $|C|$ 2-cliques $X_c = \{v_c, v\}$. Construct a new graph from $G_v$ as follows. Introduce a node $\mathbf{y}_{c,j}$ for each clique $X_c$ for each of the $j = 0, \ldots, |v| - 1$ configurations of $v$ for the settings $v_c = 0$. Connect all nodes pairwise if they correspond to different configurations of $v$. The resulting graph is a complete $|v|$-partite graph which is known to be perfect (Berge & Chvátal, 1984). To obtain the NMRF from the current complete $|v|$-partite graph, sequentially introduce additional nodes $\mathbf{y}_{c,i|v|+j}$ for each $X_c$ for each of the $j = 0, \ldots, |v| - 1$ configurations of $v$ as well as for each of the remaining $i = 1, \ldots, |v_c| - 1$ settings of $v_c$. Each sequentially introduced node is connected to the corresponding node $\mathbf{y}_{c,0+j}$ that is already in the graph as well as all its neighbors. By Lemma 2, this sequential introduction of additional nodes and edges maintains graph perfection. Once all nodes are added, the resulting graph is precisely the graph $\mathcal{G}_v$ obtained by converting a star graph $G_v$ into its equivalent NMRF form. Therefore, $\mathcal{G}_v$ is perfect. Applying induction on the star graph extends the perfect graph argument to the more general case where $G$ is a tree. Consider two star graphs: the first star $G_v$ contains nodes $\{v_1, \ldots, v_{|C|}\}$ with internal node $v$ and the second star $G_w$ contains nodes $\{w_1, \ldots, w_{|D|}\}$ with internal node $w$. Consider merging these two stars by merging node $v_1$ with node $w$, merging node $w_1$ with node $v$ and merging edge $\{w, w_1\}$ and edge $\{v, v_1\}$ into a single edge $\{v, w\}$. Clearly, the resulting merged graph, denoted $G_{v+w}$ forms a tree. The stars $G_v$ and $G_w$ separately give rise to NMRFs $\mathcal{G}_v$ and $\mathcal{G}_w$ which have already been shown to be perfect. The tree $G_{v+w}$ gives rise to an NMRF denoted $\mathcal{G}_{v+w}$. Since $V(G_v) \cap V(G_w) = \{v, w\}$, it is clear that the isolated NMRFs overlap only over the configuration nodes for the edge $\{v, w\}$. Consequently, the vertices $V(\mathcal{G}_v) \cap V(\mathcal{G}_w)$ form a fully-connected clique in $\mathcal{G}_v$, in $\mathcal{G}_w$ and in $\mathcal{G}_{v+w}$. Therefore, the merged NMRF $\mathcal{G}_{v+w}$ introduces no additional cycles beyond the ones in $\mathcal{G}_v$ and $\mathcal{G}_w$ in isolation. This gluing of NMRF graphs on cliques is a special case of Chvátal's skew-partition decomposition which is known to preserve graph perfection (Chudnovsky et al., 2006). Since $\mathcal{G}_v$ and $\mathcal{G}_w$ are perfect graphs, $\mathcal{G}_{v+w}$ must then also be a perfect graph. By induction, merging additional stars in this manner to sequentially construct any tree $G$ produces an NMRF $\mathcal{G}$ which must be a perfect graph.*

Next consider the case where the graphical model $G$ corresponds to a maximum weight bipartite matching problem (Bayati et al., 2005; Huang & Jebara, 2007; Sanghavi et al., 2008) which is known to produce integral linear programming relaxations.

**Lemma 4** *The LP relaxation of the graphical model for maximum weight bipartite matching*

$$p(X) = \prod_{i=1}^{n} \delta\left(\sum_{j=1}^{n} x_{ij} \leq 1\right) \delta\left(\sum_{j=1}^{n} x_{ji} \leq 1\right) \prod_{k=1}^{n} e^{f_{ik} x_{ik}}$$

*with non-negative $f_{ij} \geq 0$ and binary $x_{ij}$ for all $i, j = 1, \ldots, n$ is integral and produces the MAP estimate.*

**Proof 6** *The graphical model is in NMRF form so $G$ and $\mathcal{G}$ are equivalent. $\mathcal{G}$ is the line graph of a (complete) bipartite graph (i.e. a Rook's graph). Therefore, $\mathcal{G}$ is perfect, the LP is integral and recovers the MAP estimate via Theorem 2.*

A generalization of the bipartite matching problem is the unipartite matching problem. It is known that the standard LP relaxation for such problems is not always integral[7]. However, (Sanghavi et al., 2008) shows that belief propagation produces the MAP estimate in the unipartite case if the LP relaxation is integral. It is now possible to show when the LP is integral by recognizing perfect graphs and guaranteeing the convergence of belief propagation a priori.

**Lemma 5** *The LP relaxation of the graphical model $G = (V, E)$ for maximum weight unipartite matching*

$$p(X) = \prod_{i \in V} \delta\left(\sum_{j \in Ne(i)}^{n} x_{ij} \leq 1\right) \prod_{ij \in E} e^{f_{ij} x_{ij}}$$

---

[7]The nonintegrality of the LP in unipartite matching is why additional Blossom inequalities constraints are imposed in Edmonds' algorithm (Edmonds, 1965). To ensure integrality for any graph, one introduces an exponential number of Blossom inequalities: for every set of edges between an odd sized set of vertices and the remaining vertices, the sum over the set of edge weights is at least 1.

with non-negative $f_{ij} \geq 0$ and binary $x_{ij}$ for all $ij \in E$ is integral and produces the MAP estimate if $G$ is a perfect graph.

**Proof 7** *The graphical model is in NMRF form and graphs $G$ and $\mathcal{G}$ are equivalent. By Theorem 2, the LP relaxation is integral and recovers the MAP estimate if $\mathcal{G}$ is a perfect graph.*

## 6 PRUNING NMRFs

Clearly, as was the case in the previous two lemmas, if the original graphical model $G$ has some clique functions that are already nand functions (as in matching problems), then re-expanding these into NMRFs by the method in Section 3 is redundant. Therefore, only when the variables are involved in clique functions that are not nand-structured, should the conversion from $X_c$ to $\mathbf{X_c}$ be implemented.

In addition, we provide the following two procedures which are useful for pruning the NMRF prior to verifying perfection of the graph and/or MAP estimation. The procedures are DISCONNECT and MERGE. We emphasize that these can be applied to $\mathcal{G}$ optionally. Both are efficient and may simplify the NMRF hopefully converting an otherwise imperfect graph NMRF into an equivalent perfect graph NMRF (for example by exploiting additional structure in the values of the clique functions) thereby allowing exact MAP estimation. Also, the subsequent perfect graph recognition algorithm and MAP linear program can only be sped up by these procedures.

First, we obtain a graph DISCONNECT($\mathcal{G}$) from $\mathcal{G}$ by applying the DISCONNECT procedure to all nodes in the NMRF that correspond to the minimal configurations of each clique $\psi_c(X_c)$. In other words, for each $c \in C$, denote the minimal configurations of $c$ as the set of nodes $\{\mathbf{x}_{c,k}\}$ such that $f_{c,k} = \min_\kappa f_{c,\kappa} = \log(1+\epsilon)$. DISCONNECT removes the edges between these nodes and all other nodes in the clique $\mathbf{X}_c$. This is because the minimal configurations, if asserted (set to unity) or otherwise, will have no significant effect on the MAP score. Therefore, if they violate the nand relationship with other variables in $\Psi_c(\mathbf{X}_c)$ and are set to unity in addition to the other variables in $\mathbf{X}_c$, an equivalent MAP estimate can be found by setting these variables to zero while preserving a MAP estimate. In other words, given the MAP $\mathbf{X}^*$ estimate of $\rho(\mathbf{X})$ in the graph DISCONNECT($\mathcal{G}$), if more than one setting in $\mathbf{X}_c^*$ is active, only the maximal setting is preserved as a post-processing. Since minimal configurations are allowed to be redundantly asserted by the DISCONNECT procedure and may conflict with the true assignment, these are set to zero by a final post processing procedure. After MAP estimation, given all asserted variables in $\mathbf{X}_c^*$, only one $\mathbf{x}_{c,k}$ is kept asserted: the one which corresponds to the largest $f_{c,k}$ and all others which have $f_{c,k} = \log(1 + \epsilon)$ get set to zero. This does not change the score of the MAP estimate. The DISCONNECT procedure only requires $O(|\mathsf{V}(\mathcal{G})|)$.

Second, we apply another procedure to the current NMRF which is called MERGE. This procedure returns a graph where nodes in the input graph are merged. For any pair of disconnected nodes $\mathbf{x}_{c,k}$ and $\mathbf{x}_{d,l}$ in the NMRF that have the same connectivity to the rest of the graph $\mathsf{Ne}(\mathbf{x}_{c,k}) = \mathsf{Ne}(\mathbf{x}_{d,l})$, MERGE combines them into a single equivalent variable $\mathbf{x}_{c,k}$ with the same connectivity and updates its corresponding weight as $f_{c,k} \leftarrow f_{c,k} + f_{d,l}$. Then, following MAP estimation, the setting for $\mathbf{x}_{d,l}$ is recovered simply by setting it to the value of $\mathbf{x}_{c,k}$. It is straightforward to see that the procedure MERGE requires no more than $O(|\mathsf{V}(\mathcal{G})|^3)$. Thus, once the NMRF $\mathcal{G}$ is obtained via Section 3, we obtain $\mathcal{G}' = \text{MERGE}(\text{DISCONNECT}(\mathcal{G}))$ which potentially can be more readily tested for perfection and admits more efficient MAP estimation due to the simplification of the graph. Given the MAP estimate from $\mathcal{G}'$, it is straightforward to recover the MAP estimate for $\mathcal{G}$ and then reconstruct the MAP estimate of $G$.

## 7 MESSAGE PASSING

While linear programming can be used to solve for the MAP configuration whenever the NMRF involves a perfect graph, a faster approach is to perform message passing since such algorithms exploit the sparse graph topology more directly. Guarantees for the exactness and convergence of max-product belief propagation are known in the case of singly linked graphs, junction trees, single loop graphs and matching problems (Wainwright & Jordan, 2008). A more convergent algorithm was recently proposed in (Globerson & Jaakkola, 2007) which is known as convergent message passing. For binary MAP problems, it recovers the solution to the LP relaxation. It is thus investigated here as a natural competitor to linear programming for MAP estimation on the NMRF. To apply this method to an NMRF with graph $\mathcal{G} = (\mathcal{V}, \mathcal{E})$, it helps to rewrite the objective as follows:

$$\log \rho(\mathbf{X}) = \sum_{ij \in \mathcal{E}} \theta_{ij}(\mathbf{x}_i, \mathbf{x}_j).$$

Here we have defined the following potential functions:

$$\theta_{ij}(\mathbf{x}_i, \mathbf{x}_j) = \frac{\mathbf{x}_i f_i}{|\mathsf{Ne}(i)|} + \frac{\mathbf{x}_j f_j}{|\mathsf{Ne}(j)|} + \log \delta(\mathbf{x}_i + \mathbf{x}_j \leq 1)$$

where $\mathsf{Ne}(i)$ indicates all neighbors of the node $i$. Thus, all clique functions for an NMRF have the form

$$\theta_{ij}(\mathbf{x}_i, \mathbf{x}_j) = \begin{array}{c|c|c} & \mathbf{x}_j = 0 & \mathbf{x}_j = 1 \\ \hline \mathbf{x}_i = 0 & 0 & \frac{f_j}{|\mathsf{Ne}(j)|} \\ \hline \mathbf{x}_i = 1 & \frac{f_i}{|\mathsf{Ne}(i)|} & -\infty \end{array}$$

and, to avoid numerical problems, each value of $-\infty$ should be replaced with a large negative constant in practice. The convergent message passing algorithm is outlined below.

---

CONVERGENT MESSAGE PASSING:
Input: Graph $\mathcal{G} = (\mathcal{V}, \mathcal{E})$ and $\theta_{ij}$ for $ij \in \mathcal{E}$.
1. Initialize all messages to any value.
2. For each $ij \in \mathcal{E}$, simultaneously update
$\lambda_{ji}(\mathbf{x}_i) \leftarrow -\frac{1}{2} \sum_{k \in \mathsf{Ne}(i) \setminus j} \lambda_{ki}(\mathbf{x}_i)$
$\qquad + \frac{1}{2} \max_{\mathbf{x}_j} \left[ \sum_{k \in \mathsf{Ne}(j) \setminus i} \lambda_{kj}(\mathbf{x}_j) + \theta_{ij}(\mathbf{x}_i, \mathbf{x}_j) \right]$
$\lambda_{ij}(\mathbf{x}_j) \leftarrow -\frac{1}{2} \sum_{k \in \mathsf{Ne}(j) \setminus i} \lambda_{kj}(\mathbf{x}_j)$
$\qquad + \frac{1}{2} \max_{\mathbf{x}_i} \left[ \sum_{k \in \mathsf{Ne}(i) \setminus j} \lambda_{ki}(\mathbf{x}_i) + \theta_{ij}(\mathbf{x}_i, \mathbf{x}_j) \right]$
3. Repeat 2 until convergence.
4. Find $b(\mathbf{x}_i) = \sum_{j \in \mathsf{Ne}(i)} \lambda_{ji}(\mathbf{x}_i)$ for all $i \in \mathcal{V}$.
5. Output $\hat{\mathbf{x}}_i = \arg\max_{\mathbf{x}_i} b(\mathbf{x}_i)$ for all $i \in \mathcal{V}$.

---

The algorithm iterates until convergence and produces the approximate solution denoted $\hat{\mathbf{X}} = \{\hat{\mathbf{x}}_1, \ldots, \hat{\mathbf{x}}_N\}$. A key property of the algorithm is that it recovers the same solution as the LP when the variables are binary.

**Theorem 5** *(Globerson & Jaakkola, 2007) With binary variables $\mathbf{x}_i$, fixed points of convergent message passing recover the optimum of the LP.*

Thus, for binary problems, instead of solving the LP, it is possible to simply run message passing. We previously showed that when the graph $\mathcal{G}$ is a perfect graph the LP is integral and thus, in such settings, message passing recovers the MAP assignment. This permits the following corollary.

**Corollary 3** *Convergent message passing on an NMRF with a perfect graph finds the MAP estimate.*

The above thus generalizes the possible settings in which message passing converges to the MAP estimate from singly linked graphs, single loop graphs and matching graphs to the broader set of perfect graphs.

## 8 EXPERIMENTS

To evaluate the optimality of message passing, we investigate convergence on the following basic Berge graphs: bipartite graphs, complements of bipartite graphs, line graphs of bipartite graphs and complements of these line graphs. We also consider arbitrary random graphs which may or may not be perfect. Message passing was used to solve unipartite matching as in Lemma 5 on these graphs with random edge weights sampled uniformly between $[0, 1]$. To show convergence to the MAP problem, the message passing (i.e. LP) estimate is compared to the exact solution using Edmonds' algorithm. Figure 2 show the scores obtained by message passing on the vertical axis and by exact MAP estimation on the horizontal axis. Clearly, the four subfamilies of perfect graphs obtained the MAP estimate via message passing (or LP) while suboptimal solutions were recovered on arbitrary graphs (which need not be perfect).

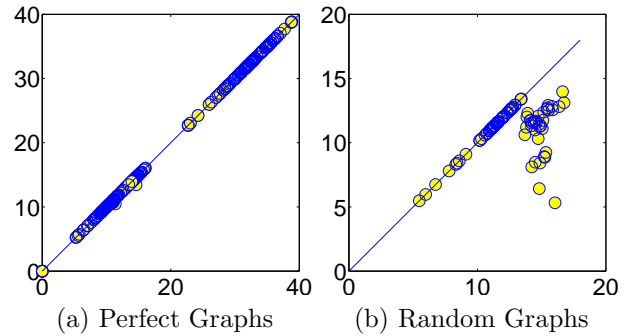

(a) Perfect Graphs    (b) Random Graphs

Figure 2: Scores for the exact MAP estimate (horizontal axis) and message passing estimate (vertical axis) for random graphs and weights. Figure (a) shows scores for four types of basic Berge graphs while (b) shows scores for arbitrary graphs. Minor score discrepancies on Berge graphs arose due to numerical issues and early stopping.

## 9 DISCUSSION

A procedure was provided to convert any graphical model into a nand Markov random field. The NMRF graph can then be efficiently diagnosed to determine if it is perfect. If it (or a pruned version of the NMRF) is perfect, MAP estimation is in P and can be solved efficiently via linear programming (or via message passing). This extends MAP estimation and message passing guarantees to a wider range of graphical models.

If the resulting NMRF is not a perfect graph, it may be useful to explore slight modifications to the MAP problem to produce a perfect graph. Replacing otherwise intractable MAP estimation with exact MAP estimation (via linear programming or message passing) on a surrogate problem is a direction of ongoing interest. Furthermore, due to the particular nature of nand Markov random fields, it may be the case that simpler variants of message passing (for instance, its predecessor, the max product algorithm (Globerson & Jaakkola, 2007)) may also have convergence guaran-tees. One direction for future work is the conjecture that max product on NMRFs with perfect graphs also recovers the MAP estimate.

## 10 ACKNOWLEDGMENTS

The author thanks M. Chudnovsky and D. Dueck for discussions and the anonymous referees for helpful suggestions.